\documentclass{./llncs2e/llncs}
\usepackage{graphicx}
\usepackage{fixltx2e}
\usepackage{mathtools}
\usepackage{bm}
\usepackage[nolist,nohyperlinks]{acronym}
\usepackage[section]{placeins}

\usepackage{float}

\usepackage[scientific-notation=true, round-precision=2, round-mode=figures]{siunitx}

\usepackage{footnote}
\usepackage{booktabs}
\usepackage{longtable}
\usepackage{babel}
\usepackage{{threeparttable}}

\usepackage{textcomp}

\usepackage{caption}
\usepackage[utf8]{inputenc}

\usepackage[acronym]{glossaries}

\newacronym{MA}{MA}{My Acronym}

\makeglossaries

\def \Title{An End to End Network Architecture for Fundamental Matrix Estimation}

\def \Author{Yesheng Zhang, Xu Zhao, Dahong Qian}

\def \Institution{Shanghai JiaoTong University}

\begin{document}

\title{\Title}
\author{\Author}
\institute{\Institution}

\maketitle



\begin{abstract}
  In this paper, we present a novel end-to-end network architecture to estimate fundamental matrix directly from stereo images. To establish a complete working pipeline, different deep neural networks in charge of finding correspondences in images, performing outlier rejection and calculating fundamental matrix, are integrated into an end-to-end network architecture.
 To well train the network and preserve geometry properties of fundamental matrix, a new loss function is introduced. To evaluate the accuracy of estimated fundamental matrix more reasonably, we design a new evaluation metric which is highly consistent with visualization result. Experiments conducted on both outdoor and indoor data-sets show that this network outperforms traditional methods as well as previous deep learning based methods on various metrics and achieves significant performance improvements.
\end{abstract}



\section{Introduction}
    Fundamental matrix is a critical concept in geometric-based computer vision and it contains a complete description of the projective structure of a set of two cameras \cite{Luong:1996kc}. So that the estimation of fundamental matrix becomes an important computer vision task which is essential to many applications like camera calibration \cite{Faugeras:1992dn,Boukamcha:2019tj}, images rectification \cite{Mallon:bx},  robotic visual servoing \cite{Engineering:ck} and so forth.
    
    Fundamental matrix estimation is a long-explored problem \cite{Hartley:2003un}, but for many real world applications, it's still stucked in two main difficulties, namely, feature correspondence and model estimation. As a typical example, the well-established traditional fundamental matrix estimation pipeline is consisted of three main modules, SIFT \cite{international:fa}, RANSAC \cite{Fischler:1987bf} and 8-points algorithm \cite{Hartley:2003un}. SIFT, or other interesting point feature extractors are used to detect features and establish correspondences, while RANSAC or 8-point algorithm is used to perform model estimation. This pipeline, however, is prone to be unreliable when images are lack of features or correspondence errors occur. RANSAC and 8-point algorithm (or 7-point algorithm \cite{Hartley:2003un}), as separated modules from the feature correspondence part, are also not robust enough to well handle the observatory errors introduced during extracting features. The noise and incorrect matching inherited from inaccurate correspondences could lead to huge estimation error.     
    
    With the advent of deep learning, many related methods based on deep neural networks are presented and show potentials to this task. And these methods can be divided into two categories: two-stage and one-stage methods. Two-stage methods combine deep neural network and traditional algorithms to construct the whole pipeline for fundamental matrix estimation. In \cite{Choy:2016up}, CNN-based network is used to establish correspondences between two images, and fundamental matrix is calculated by RANSAC and 8-point algorithm with these correspondences. In \cite{Zhang:2019ub}, to perform outlier rejection, PointNet\cite{Qi:2017vq}-based network is used to get the weights of each correspondence which describe the matching accuracy by defining outlier rejection as classification task. It outperforms traditional RANSAC-like methods and can be combined with SIFT-like algorithm and 8-point algorithm to estimate the fundamental matrix. While these methods outperform traditional algorithms on standard benchmarks, incorporating them into the classic pipeline may not make sure performance increase. For example, if the SIFT-like algorithm failed due to bad image quality, the PointNet-based method can not select the inliers correctly and the overall performance could be poor. This inconsistency is also reported in \cite{bian2019evaluation}. 
    
    On the other hand, one-stage methods stand for the end-to-end network architecture to this task. In \cite{Poursaeed:2018vh}, feature correspondences built by CNN-based network and fully connected layers are integrated to estimate the fundamental matrix. This network advances this task one more big step by establishing an end-to-end framework. But only simple fully-connect layers are not enough to well handle the matching errors and achieve overall satisfactory performance. Comparing with traditional pipeline, this architecture is also in absence of outlier rejection mechanism, which is critical to the final performance.
    To forge a better one-stage method, in this work, we propose an end-to-end deep neural network architecture for fundamental matrix estimation, which uses CNN-based network to establish feature correspondence and PointNet-based network combined with fully-connected layers to perform model estimation. It's trained with 
   end-to-end manner, so the overall accuracy of fundamental matrix estimation is successfully improved. Furthermore, we also define a novel loss function, which can be used to describe the geometry property of a fundamental matrix to guide the network training so the performance can be increased efficiently. Our work is evaluated on both outdoor and indoor datasets and achieves significant accuracy improvements.
    
    To evaluate the estimated fundamental matrix, usually some commonly-used metrics, like epipolar constraint and symmetric epipolar distance, are employed. When evaluating the accuracy of fundamental matrix, however, these metrics only focus on the relationship between points and corresponding epipolar lines. But by visualization during the experiments, we find that the orientation of epipolar lines may deviates from ground truth when the points are lied on lines correctly. That means the metrics mentioned above can't measure the error of epipolar lines' orientation caused by the inaccuracy of fundamental matrix. So we design a new metric, by which the accuracy can be evaluated by the angle between lines drawn from estimated fundamental matrix and ground truth respectively. So the evaluation could be more comprehensive.
    
    In sum, the main contributions of our work are threefold:
    \begin{enumerate}
    	\item We propose an end-to-end deep neural network architecture with completed pipeline from feature correspondence to model estimation to get fundamental matrix from stereo images directly.
    	\item A novel loss function is designed to guide the network training by preserving the geometry property of fundamental matrix and is prone to be efficient on performance improvement.
    	\item A novel metric which measures the orientation error of epipolar lines drawn by estimated fundamental matrix is presented to evaluate the fundamental matrix comprehensively.
    \end{enumerate}


\section{Related Work}
The epipolar constraint is a basic constraint when considering the case that two cameras look at the same scene \cite{Zhang:1998il}. For each point $\bm{m}=[x, y]$ in the first camera, its correspondence $\bm{m'}=[x', y']$ lies on its epipolar line ${l'}_{\bm{m}}$. This correspondence between points and lines is encoded in a $3 \times 3$ matrix {\it F} called fundamental matrix \cite{Luong:1996kc} with rank 2. It can be formulated as

        \begin{equation}
            l'_{\bm{m}} = F\bm{m}
        \end{equation}

As the point lies on the epipolar line, it follows that

        \begin{equation}
            {\bm{m'}}^T F \bm{m} = 0
        \end{equation}
This is one of the most important geometric properties of fundamental matrix called {\it epipolar constraint}. The fundamental matrix can be computed as

        \begin{equation}
            F = K_2^{-T}[t]_{\times}RK_1^{-1}
        \end{equation}
where $K_1, K_2$ are camera intrinsic matrix, and $R, t$ represent the relative camera rotation and translation respectively. This indicates that the fundamental matrix only depends on the cameras’ internal parameters and their relative pose. So if one wants to get camera pose only from image measurements, the Fundamental matrix is the key concept\cite{Luong:1996kc}.

\subsection{Traditional Methods} 
For decades, traditional fundamental matrix estimation methods have a well-established pipeline including feature correspondence and model estimation. Regarding to feature correspondence, they use SIFT-like algorithms to detect and establish putative correspondences. However, these algorithms fail when images are lack of texture or blurred. 

After correspondence establishment, the methods for model estimation can be classified into linear, iterative and robust methods\cite{Armangue:uh}. Linear and iterative methods are all focused on optimization questions,  and they have different criterions like:
        
        \begin{equation}
            \min_F \sum_i ({\bm{m'}}^T F \bm{m})^2
        \end{equation}

And the classic linear method which is called 8-point method is based on this minimization equation.

{ \it The symmetric epipolar distance} which describes the geometric distance between points and corresponding epipolar lines is also used as criterion on some iterative methods:

        \begin{equation}
            \min_F \sum_i (\frac{1}{(F\bm{m}_i)^2 _1 + (F\bm{m}_i)^2 _2} + \frac{1}{(F\bm{m}_i')^2 _1 + (F\bm{m}_i')^2 _2})({\bm{m}'}_i^T F \bm{m}_i)^2
            \label{Cr:SED}
        \end{equation}

However, linear and iterative methods cannot hold with correspondence outliers and may result in big errors\cite{Armangue:uh}. 
        So it's necessary to deal with outliers by outlier rejection which needs robust methods.

The most widely used approach to perform outlier rejection is RANdom SAmple Consensus (RANSAC)\cite{Fischler:1987bf,Lacey:vg} which samples randomly to search a geometric model that has the most support in the form of inliers. And there are a large amount of algorithms inherit this basic idea with some variations like MLESAC\cite{Luong:1996kc} (Maximum LikElihood SAmple and MAPSAC\cite{Torr:2002jb}(Maximum A Posteriori SAmple Consensus). MLESAC searches the best model by maximizing a likelihood with the same point selection strategy as the RANSAC. MAPSAC includes Bayesian probabilities in minimization to be more robust against noise and outliers.

Although the robust methods can cope with noise in the location of points and false matching, they still can be unreliable by modeling a wrong geometry\cite{Hartley:2003un}. 
\subsection{Neuron Network-Based Methods}

Different with traditional methods, neural network-based methods focus on feature correspondence and model estimation separately. That is to say, these methods can be divided into CNN-based methods to find feature correspondence and PointNet-based methods for model estimation.

With the convolutional neural network (CNN) getting breakthrough in many computer vision tasks, CNN-based methods have been widely used in multi-view geometry and 3D vision tasks. A Multi-Scale deep network is used to predict depth map from a single image in \cite{Eigen:2014vq}. \cite{Melekhov:2017to} uses CNN based approach for relative pose between two cameras and \cite{Ummenhofer:2017uq} trained a convolutional network to compute depth and camera motion simultaneously. \cite{DeTone:vr,Nguyen:gw} use deep CNN for estimating homography and \cite{:ue} recovers homography from the perspectively distorted document images by using CNN.

Regarding feature correspondence, researchers\cite{DeTone:2018tv,DeTone:2018tp,Luo:2018wq,Luo:2019ub,Li:ie,Rocco:uc}attemped to use CNN-based method to accomplish geometric matching task. \cite{Choy:2016up} proposes a deep learning framework for visual correspondence. And \cite{Poursaeed:2018vh} uses this framework and fully connect layers to estimate fundamental matrix from stereo images directly. However, the simple fullyconnected layer can hardly deal with the error in features.

Meanwhile, some works\cite{Yi:2017wj,Zhang:2019ub,Ranftl:2018wc} focus on replacing RANSAC-like algorithms with neuron network. They use a network called PointNet\cite{Qi:2017vq} which is designed to handle point clouds to deal with putative correspondence for outlier rejection. Through this kind of network, they get accurate correspondence and can be used to calculate fundamental matrix. Although the PointNet-based network can deal with the outliers, the error in putative correspondence still make a considerable impact on the overall performance in fundamental matrix estimation task. 

Compared with traditional pipeline, network-based methods are lake of a complete process in fundamental matrix estimation task, and the missing part have a considerable impact on overall performance.
So to establish a complete network-based pipeline, we combine CNN and PointNet together to form a novel network architecture for end-to-end fundamental matrix estimation.

\section{Network Architecture}
As mentioned before, our network takes advantage of different deep neural network architectures to establish a complete pipeline for end-to-end fundamental matrix estimation. Specifically, it uses CNN-based network to extract image features and set up putative correspondences and uses PointNet-based network to perform outlier rejection. With accurate correspondences, it calculates fundamental matrix by fully connected layers. These different deep neural network architectures are divided into three submodels. The detailed network architecture is described as followed.
  
\begin{figure}[t]
	\centering
	\includegraphics[height=40mm, width=120mm]{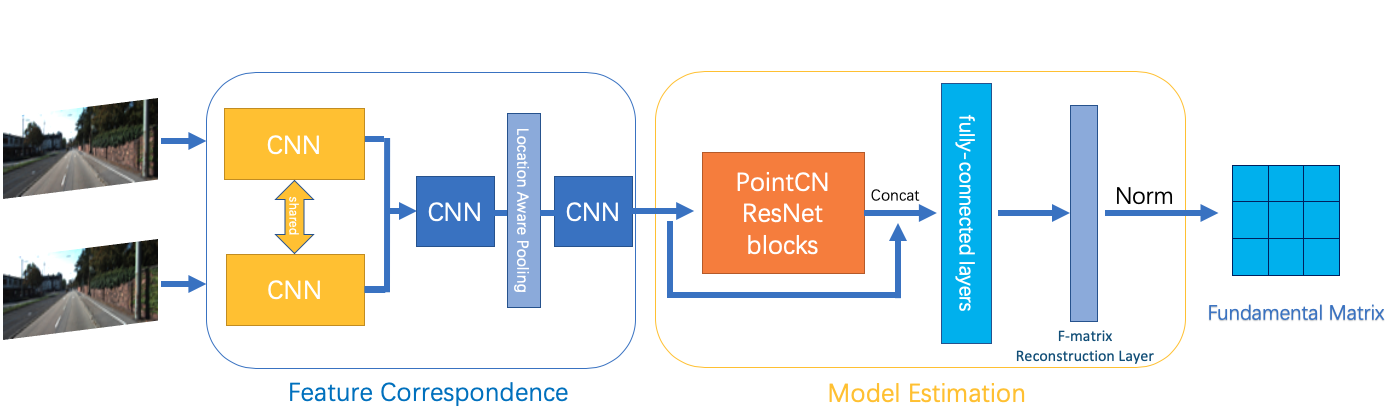}
	\caption{\small The general network architecture of our work. To establish a completed fundamental matrix estimation pipeline, we use CNN-based network({\it Submodel 1}) to perform feature correspondence and PointNet\cite{Qi:2017vq}-based network({\it Submodel 2}) along with fully-connected layers({\it Submodel 3})to perform model estimation.}
	
\end{figure}
\subsection{Putative Correspondence Establishment}

For the first part of fundamental matrix estimation pipeline, feature correspondence, we adopt a deep learning framework for accurate visual correspondences presented in \cite{Choy:2016up}(Universal Correspondence Network, UCN) to get the putative correspondences. However, the UCN framework has Spatial Transformers which could remove part of information about matching points location\cite{Choy:2016up}. So we use UCN without Spatial Transformers and add fully convolutional network after it with location aware pooling layers(LAP)\cite{Poursaeed:2018vh} which keep all the indices of where the activations come from in the maxpooling layers to get matching points position feature.

After these layers, we get the matching points position features along with image features. Then we combine them by convolution layers to establish the putative correspondences in the form of $[\bm{m},\bm{m'}]$ which can be the input features of the next part.

\begin{figure}[t]
\centering
\includegraphics[height=40mm, width=120mm]{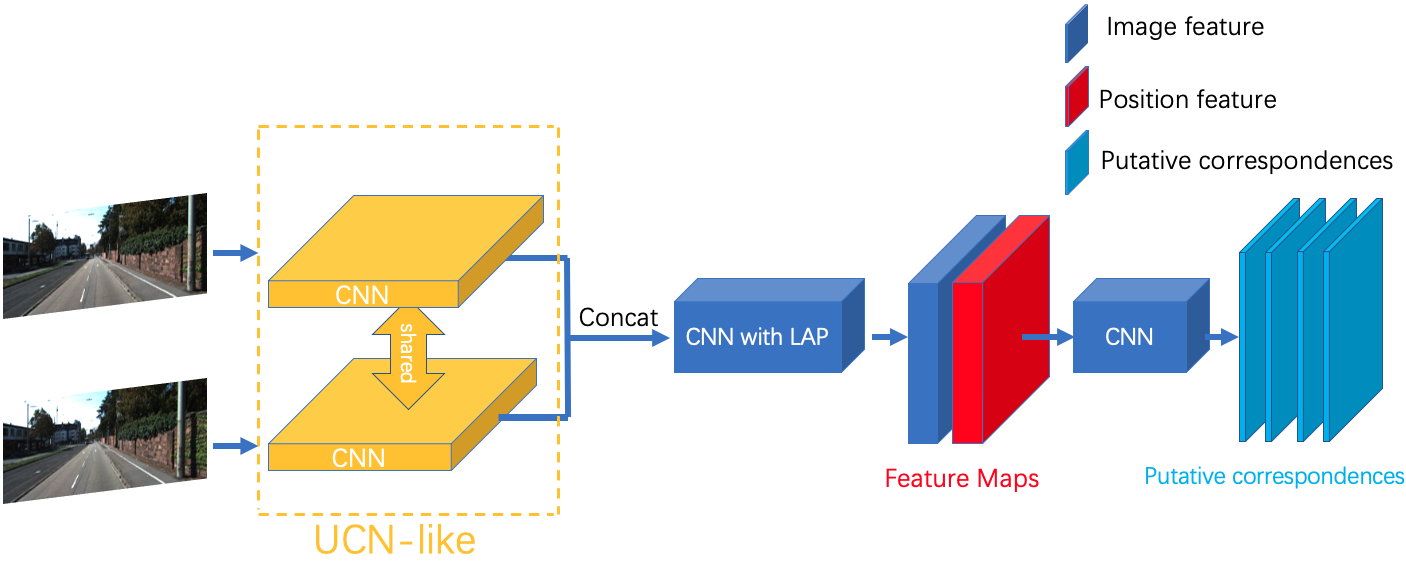}
	\caption{\small Putative Correspondence Establishment Submodel. Stereo images are passed to {\it UCN-like}\cite{Choy:2016up} part which has shared weights firstly, and then the resulting features are concatenated and passed to the {\it CNN with Location Aware Pooling(LAP)}\cite{Poursaeed:2018vh} to get the images features along with position features. Finally, we use another CNN part to get putative correspondence}
\end{figure}

\subsection{Outlier Rejection}

In \cite{Ranftl:2018wc}, PointNet-like architecture which applies Multi Layer Perceptrons (MLPs) on each point individually is used to generate weights of each putative correspondence. Those weights indicate the matching accuracy and can be used as {\it 8-point algorithm with weight} to estimate fundamental matrix. In \cite{Yi:2017wj}, PointNet is improved by adding Context Normalization(PointCN) to encode the global context which is beneficial for outlier rejection. And so we use PointCN with ResNet\cite{He:tt} parts to perform outlier rejection by generating weights of every correspondence here.

After getting weights, we concatenate it with putative correspondences as the input of fundamental matrix estimation submodel.

\begin{figure}[t]
\centering
\includegraphics[height=40mm, width=120mm]{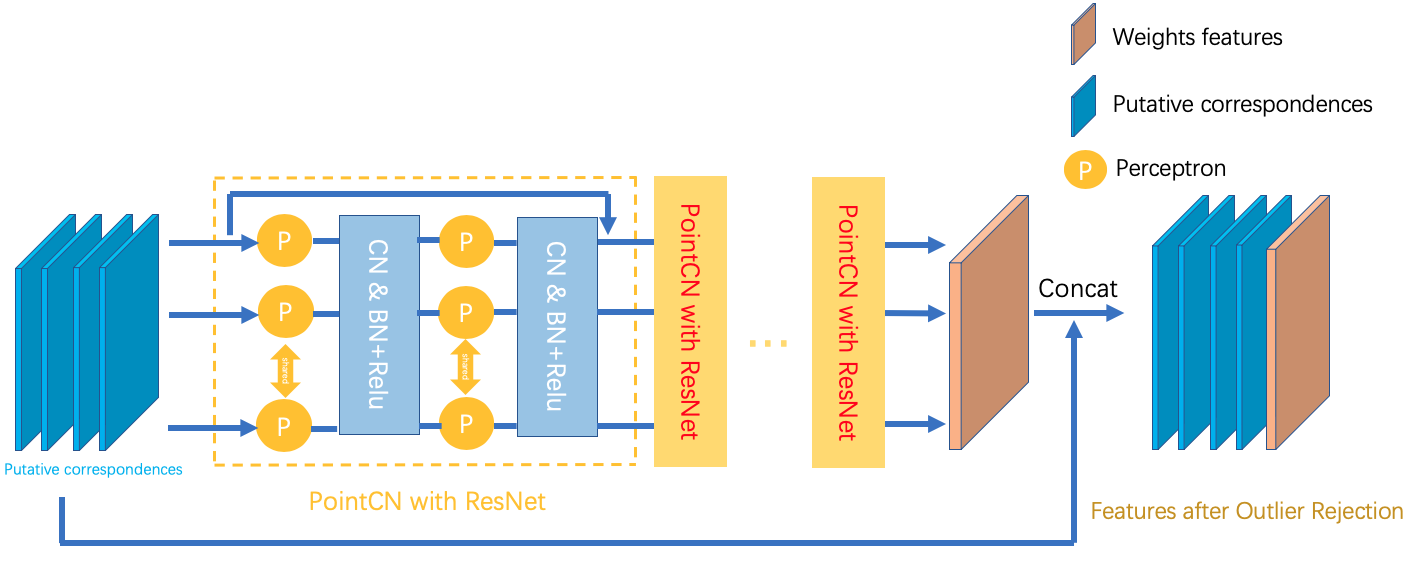}
\caption{\small Outlier Rejection Submobel. Putative Correspondences are passed to {\it PointCN with ResNet}\cite{He:tt} to perform Outlier Rejection and get the weights which represent the matching accuracy of each correspondence. After that, weights are concatenated with Putative Correspondences and fed in the next part.}
\end{figure}

\subsection{Fundamental Matrix Estimation}

With the weights and putative correspondences, the fundamental matrix estimation can be formulated as {\it 8-point algorithm with weight}:

\begin{equation}
	\min_F \sum_i w_i ({\bm{m}'}_i^TF{\bm{m}}_i)^2
\end{equation}

And this formula can be fitted by fully-connected layers.
            
The $3 \times 3$ fundamental matrix has a rank of 2 which means the first two columns $f_1,f_2$ can represent the third column $f_3$ by two coefficients $\alpha$ and $\beta$ such that $f_3=\alpha f_1 + \beta f_2 $. So we can set the output layer as $[f_1,f_2,\alpha,\beta]$ which preserve the rank of fundamental matrix which is proposed in \cite{Poursaeed:2018vh} and called F-matrix Reconstruction Layer. 

The final fundamental matrix is normalized by its maximum absolute value.

\begin{equation}
	F_{norm} = \frac{F}{\max_{i,j}|F_{i,j}|}
\end{equation}
            
\begin{figure}[t]
\centering
\includegraphics[height=40mm, width=130mm]{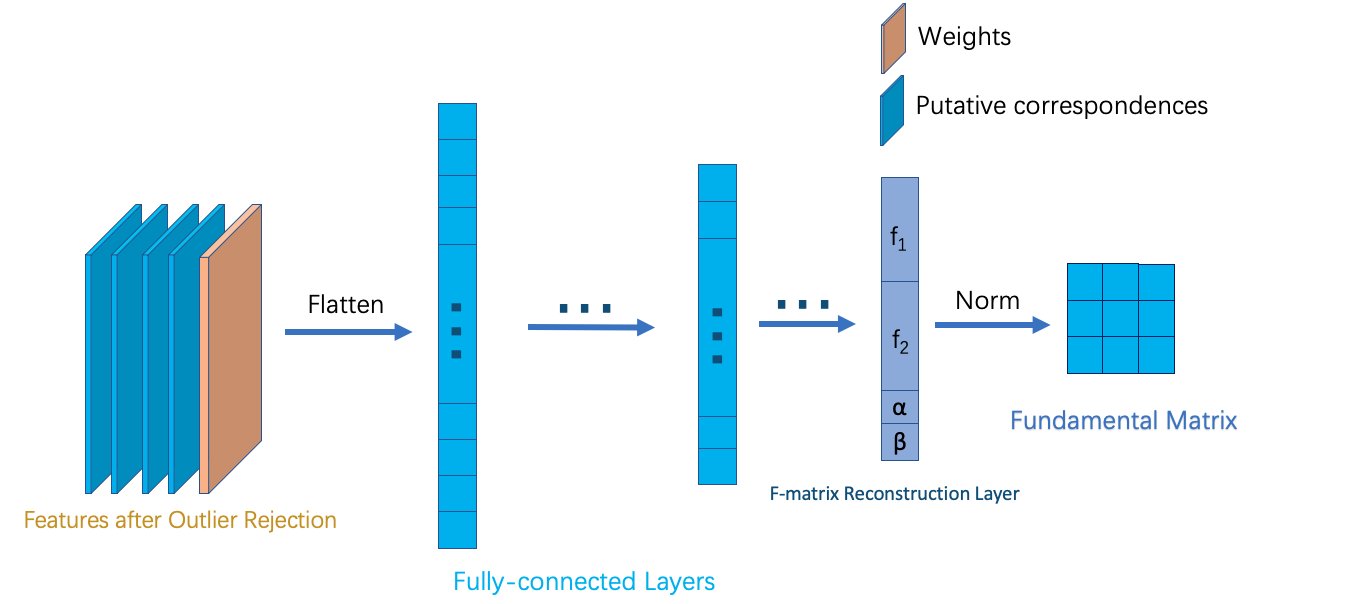}
\caption{\small Fundamental Matrix Estimation Submodel. Features after Outlier Rejection are flatten and passed to {\it fully-connected layers}. After the {\it F-matrix Reconstruction Layer}, the fundamental matrix is normalized and output.}
\end{figure}

\subsection{Model Learning}
In \cite{Poursaeed:2018vh}, simple L1, L2 loss function is used to guide the fundamental matrix estimation. However, it is clear that this loss function is not enough to describe the fundamental matrix, especially the geometry properties. 

While one of the core constraints of epipolar geometry is that the points are lied on the corresponding epipolar lines, and the corresponding epipolar line of point $m$ can be represented by fundamental matrix as $l'_{m} = Fm$. So one of the most important geometry properties of fundamental matrix is as follows which is the epipolar constraint mentioned before.

\begin{equation}
	{\bm{m}'}^T F \bm{m} = 0
\end{equation}

When we get the correct matching points, we can evaluate the accuracy of fundamental matrix through epipolar constraint. The fundamental matrix is more accurate when the epipolar constraint is closer to zero which means it can be the loss function to guide the fundamental matrix estimation.
        
So in order to keep the geometric property of fundamental matrix during the estimation, we propose a novel loss function which includes L1, L2 loss as well as epipolar constraint.

\begin{equation}
	Loss = l_{L1,L2} + l_{e}
\end{equation}

$l_{L1,L2}$ is the L1, L2 loss of the fundamental matrix compared with ground truth.

\begin{equation}
	l_{L1,L2} = \alpha \sum_{i,j}||\hat{F} - F_{GT}||_1 + \beta \sum_{i,j}||\hat{F} - F_{GT}||_2
\end{equation}

And $l_e$ is the average of residual of ${\bm{m}'}_i^TF\bm{m}_i$, where $\bm{m}_i, \bm{m}'_i$ are correct matching points(inliers) which is chosen by $F_{GT}$(Inlier Threshold is $10^{-2}$ of symmetric epipolar distance (\ref{Cr:SED}).).
            
\begin{equation}
	l_e = \gamma \frac{1}{N} \sum_i^N |{\bm{m}'}_i^T\hat{F}\bm{m}_i|
\end{equation}
            
By choosing the appropriate weights $\alpha, \beta, \gamma$, this loss function can guide the network to estimate the accurate fundamental matrix and keep its geometric property at the same time.


\section{Experiments}
To prove that our network can learn the fundamental matrix from stereo images successfully, we train our model on both outdoor and indoor data-sets and compare it with many other traditional and deep neuron network-based methods on various metrics.

\subsection{Datasets}

\subsubsection{Outdoor Scenes}
We use the KITTI data-set which provides stereo images recorded from 22 distinct driving sequences, eleven of which have publicly available ground truth odometry\cite{Geiger:2013kp}. We adopt 2000 images from the raw data in the category 'City', and divide it into 1600 images for training, 200 images for validation and 200 images for testing. Ground truth Fundamental matrix can be calculated by the ground truth camera parameters. 

\subsubsection{Indoor Scene}
We use the TUM SLAM data-set which provides videos of indoor scene. Although this data-set is not taken by stereo cameras, it is processed in \cite{bian2019evaluation} to suit the fundamental matrix estimation task by combining images pairs with ground truth fundamental matrix. We also chose 2000 images from it and split them in the same way of KITTI data-set.

\subsection{Evaluation Metrics}
The following metrics are commonly used to evaluate the estimated fundamental matrix accuracy. And all the correspondences used are inliers filtered by ground truth Fundamental matrix(Inlier Threshold is $10^{-2}$ of symmetric epipolar distance).

{\it Epipolar Constraint: }
\begin{equation}
	M_{EC}(\hat{F},\bm{m},\bm{m'}) = \frac{1}{N}\sum_i^N |{m'}_i^T F m_i| 
\end{equation}

{\it Symmetric Epipolar Distance: }
\begin{equation}
	M_{ED}(\hat{F},\bm{m},\bm{m'}) = \frac{1}{N}\sum_i^N (\frac{1}{(Fm_i)^2 _1 + (Fm_i)^2 _2} + \frac{1}{(Fm_i')^2 _1 + (Fm_i')^2 _2})({m'}_i^T F m_i)^2
\end{equation}

We also used the estimated fundamental matrix to draw epipolar lines as the visualization result which can be compared with epipolar lines drawn by the ground truth fundamental matrix.

\subsection{Novel Metric}

\begin{figure}[t]
	\centering
	\includegraphics[width=120mm,height=50mm]{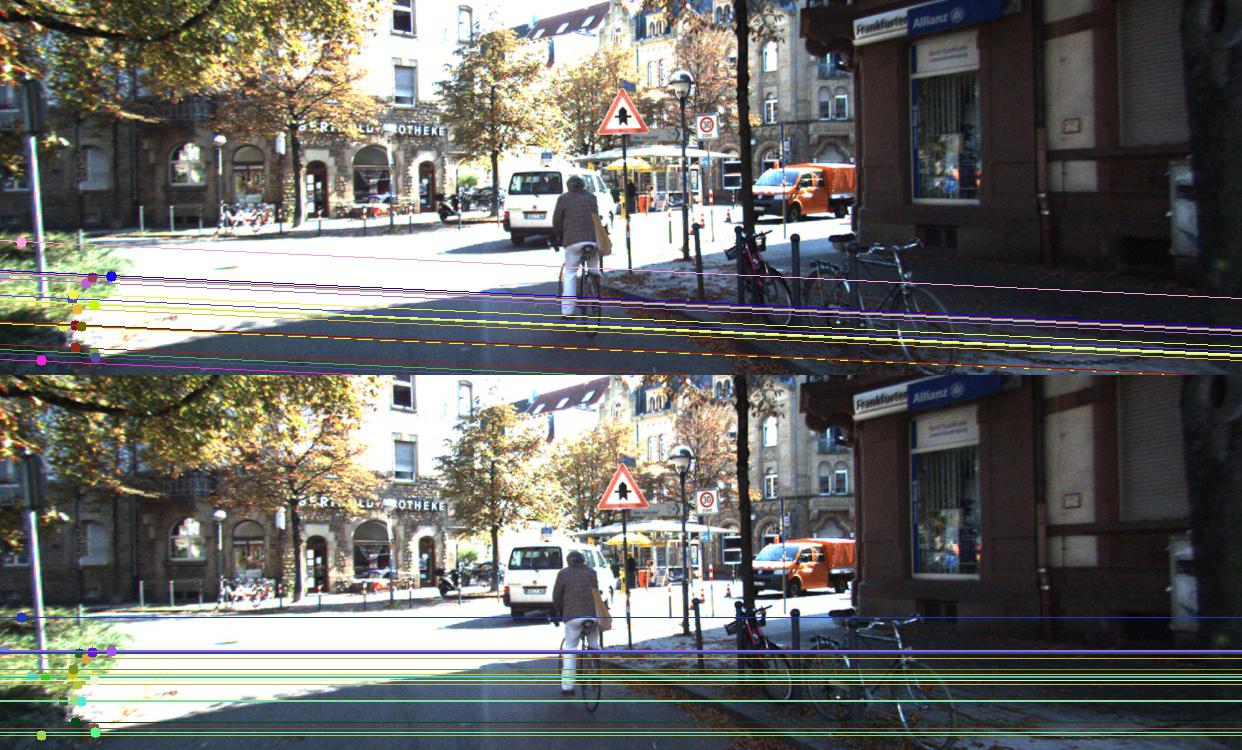}
	\caption{\small The comparison of epipolar lines drawn by fundamental matrix estimated by RANSAC(above) and the ground truth(below) one. It can be seen that the orientation of epipolar lines drawn by fundamental matrix estimated by RANSAC have deviation from ground truth when the points are lied on correspond lines. That means metrics which only check the relationship between points and epipolar lines can not measure the error of epipolar lines orientation.}
	\label{Nov:Metric}
\end{figure}

However, a problem arose during the experiments. Metrics mentioned above only check whether the points lie on the correspond epipolar lines, but errors in orientation of epipolar lines still can happen as shown on Fig. \ref{Nov:Metric}.

It is clear that epipolar lines orientation are different between RANSAC and ground truth in Fig. \ref{Nov:Metric}, but the points lie on the epipolar lines exactly which means RANSAC algorithm can achieve good performance on metrics like Epipolar Constraint or Symmetry Epipolar Distance. Thus we need a metric to measure the error of epipolar lines orientation.
 
\begin{figure}[t]
	\centering
	\includegraphics[width=110mm,height=30mm]{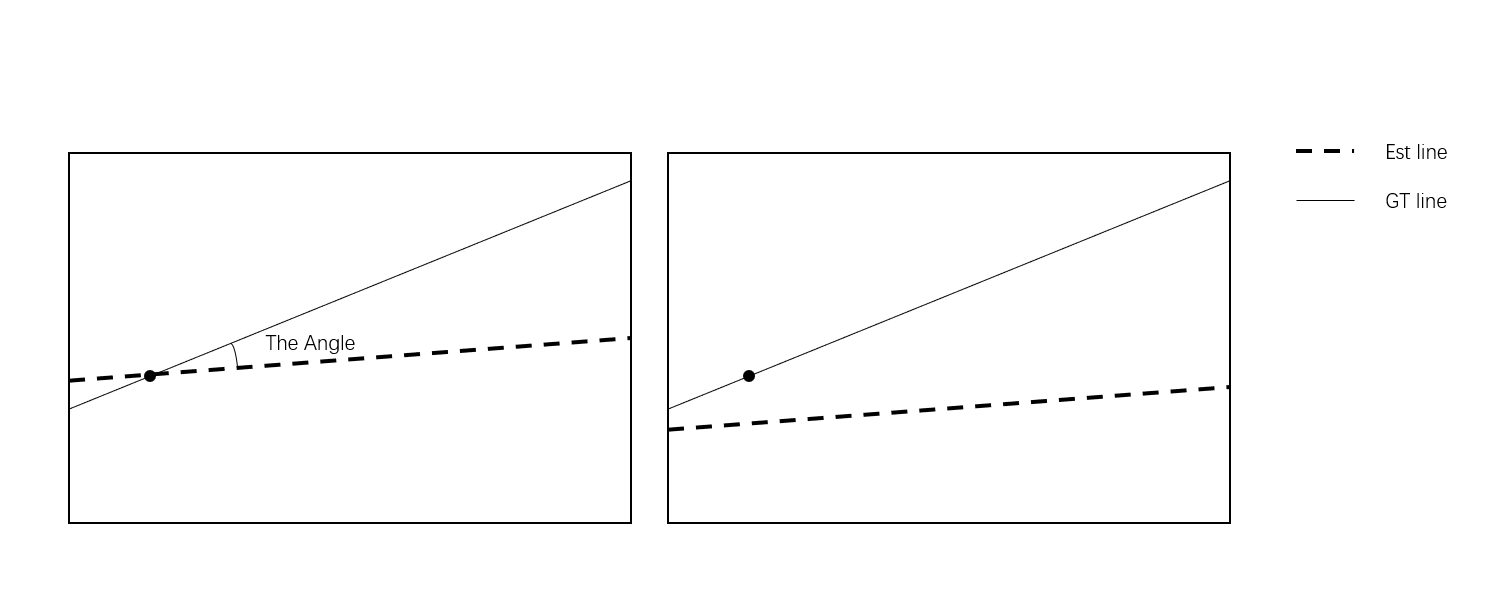}
	\caption{\small Inlier Epipolar Angle. The angle we calculate(left) and the outlier situation(right) which we don't calculate the angle between lines. In this figure, the solid line is the epipolar line drawn by ground truth Fundamental matrix({\it GT line}) and the dotted line is the epipolar line drawn by the Fundamental matrix estimated by algorithm({\it Est line}).}
     \label{Me:A}
\end{figure}

So a novel metric, as shown in Fig. \ref{Me:A}, which takes the angle between epipolar lines drawn by estimated fundamental matrix and  ground truth fundamental matrix to evaluate the accuracy of algorithms are proposed. And the intersection point of these two lines must be the corresponding point. That means if the epipolar line drawn by estimated fundamental matrix doesn't go through the corresponding point, we will take it as outlier and won't calculate the angle. So we call this metric {\it Inlier Epipolar Angle}$(M_{EA}$). 
        
To sum up, we use {\it Epipolar Constraint}($M_{EC}$), {\it Symmetry Epipolar Distance} ($M_{ED}$) and {\it Inlier Epipolar Angle}($M_{EA}$) together to evaluate the accuracy of fundamental matrix estimation.

\subsection{Implementation Details}
The Putative Correspondence Establishment submodel includes UCN-like architecture\cite{Choy:2016up} ,convolution layers and Location Aware Pooling\cite{Poursaeed:2018vh} layers. The Outlier Rejection submodel uses 6 PointCN with ResNet blocks and we concatenated its output weights with putative correspondences as the input of Fundamental Matrix Estimation submodel. This final part contains fully-connect layers and F-matrix Reconstruction Layer\cite{Poursaeed:2018vh} and outputs the normalized fundamental matrix. The parameters in loss function are $\alpha=0.1, \beta=0.01, \gamma=0.001$.

\subsection{Results and Comparison}
We selected 200 pairs of stereo images which are not used in training from KITTI raw data and TUM SLAM data-set respectively to conduct the evaluation experiments and take the average results. We also compared our algorithm with many existing algorithms. From the traditional methods, we chose SIFT algorithm to establish feature correspondences and 8-Point, RANSAC algorithm to perform model estimation. Regarding to deep neuron network-based methods, DeepF\cite{Poursaeed:2018vh} which uses CNN along with fully connect layers to estimate fundamental matrix and OANet\cite{Zhang:2019ub} which improves the PointCN in its ability to capture global and local context were chosen to conduct comparitive experiments. We used three metrics mentioned in Sec. 4.3 and choose 40 matching points every figure which are filtered by ground truth fundamental matrix(The threshold is $10^{-2}$ of Symmetric Epipolar Distance.) to calculate these metrics(N=40).

\begin{table}[!t]
      \centering
      \caption{\small Experiments result on KITTI and TUM SLAM dataset. It can be seen that our network get the best results in general on both indoor and outdoor scenes. Especially on KITTI dataset, our work improves the fundamental matrix estimation accuracy significantly. However, the accuracy gap between ground truth and our algorithm still exists on TUM SLAM dataset because of the large variation in fundamental matrix of different image pairs and the lake of texture of images.}
        \begin{tabular}{c|ccc|ccc} \hline
          Dataset & \multicolumn{3}{c}{KITTI} & \multicolumn{3}{c}{TUM SLAM} \\ \hline
          Metrics & $M_{EC}$ & $M_{ED}$ & $M_{EA}$(${}^{\circ}$) & $M_{EC}$ & $M_{ED}$ & $M_{EA}$(${}^{\circ}$)\\ \midrule
          SIFT+8Point  &   5.53 & 16364.34 & 14.56 & 10.88 & 68369.03 & 35.06\\
          SIFT+RANSAC  &   0.42 & 10.88  & 2.64 & 0.99 & 16.08 & 41.93\\
          DeepF\cite{Poursaeed:2018vh}  & 0.62 & 118.66 & 0.60 & 1.30 & 20.32 & 37.23 \\
          SIFT+OANet\cite{Zhang:2019ub} +RANSAC & 0.27 & 43.48  & 3.32 & {\it 0.61} & {9.83} & 40.61 \\
          Ours   &   {\it 0.043} & {\it 0.72} & {\it 0.45} & {0.81} & {\it 6.57} & {\it 17.33} \\ 
          GroundTruth & 0.0084 & 0.63  & 0.26 & 0.045 & 0.035 & 1.65\\ \hline
        \end{tabular}
         \label{Res:table}
\end{table}
        
The results are summarized in Tab. \ref{Res:table}. It can be seen that our algorithm got the best scores on all three metrics on KITTI data-set and are much more close to ground truth than others. While on TUM SLAM dataset, all algorithms get worse result than KITTI data-set because the texture of images are often weak and blur sometimes happen due to the fast camera movement. The bad result on $M_{EA}$ indicate the wrong model estimation which can be seen clearly in visualization results. And our algorithm get good scores in general, but the accuracy gap between ground truth and our algorithm still exists.

\begin{figure}[t]
	\centering
	\includegraphics[height=50mm, width=120mm]{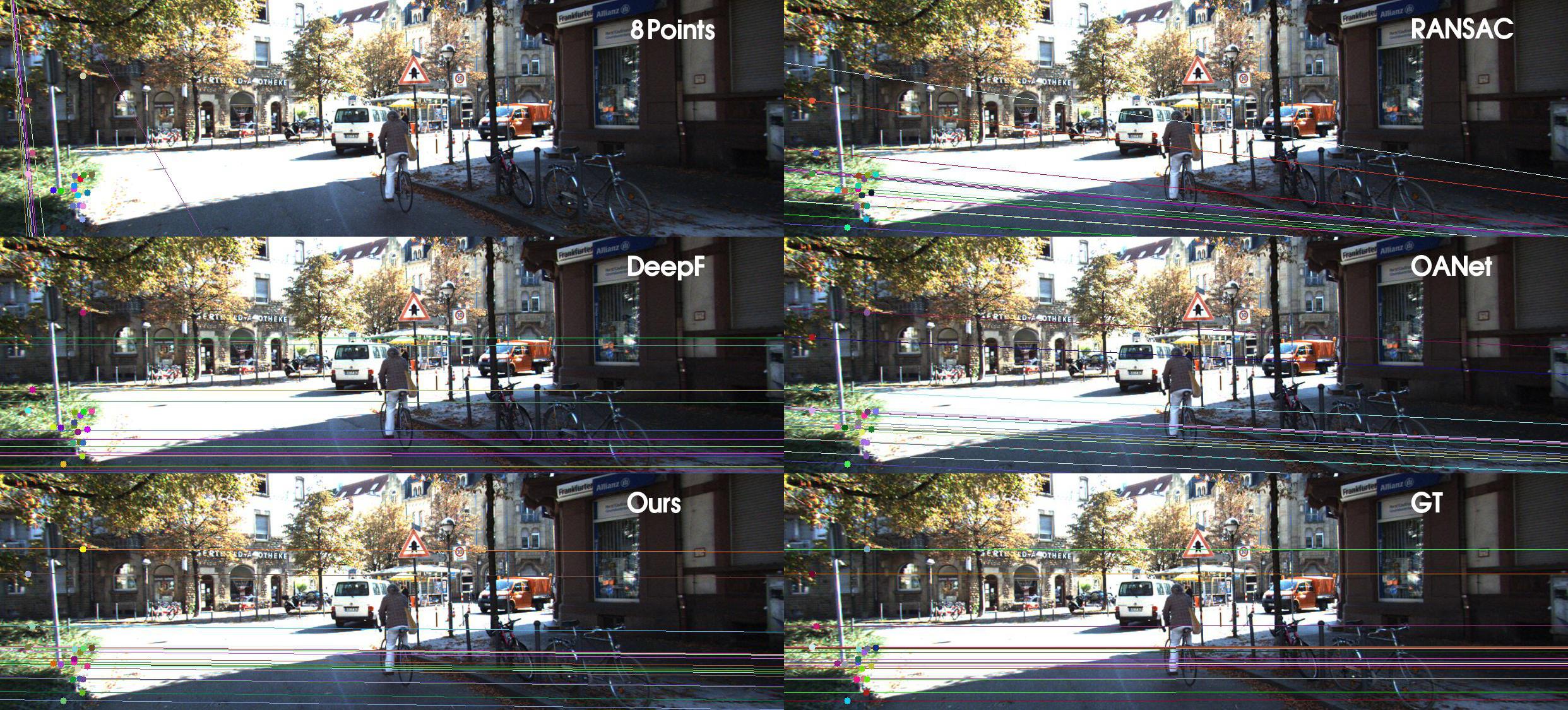}
	\caption{\small Visualization on KITTI dataset. 8-point algorithm gets the worst performance. RANSAC and OANet both have slight deviation in epipolar lines orientation. DeepF gets some points not lying on corresponding epipolar lines. Our network outperforms other algorithms and gets the closet result to ground truth.}
	\label{Vis:K}
\end{figure}

\begin{figure}[t]
	\centering
	\includegraphics[height=50mm, width=120mm]{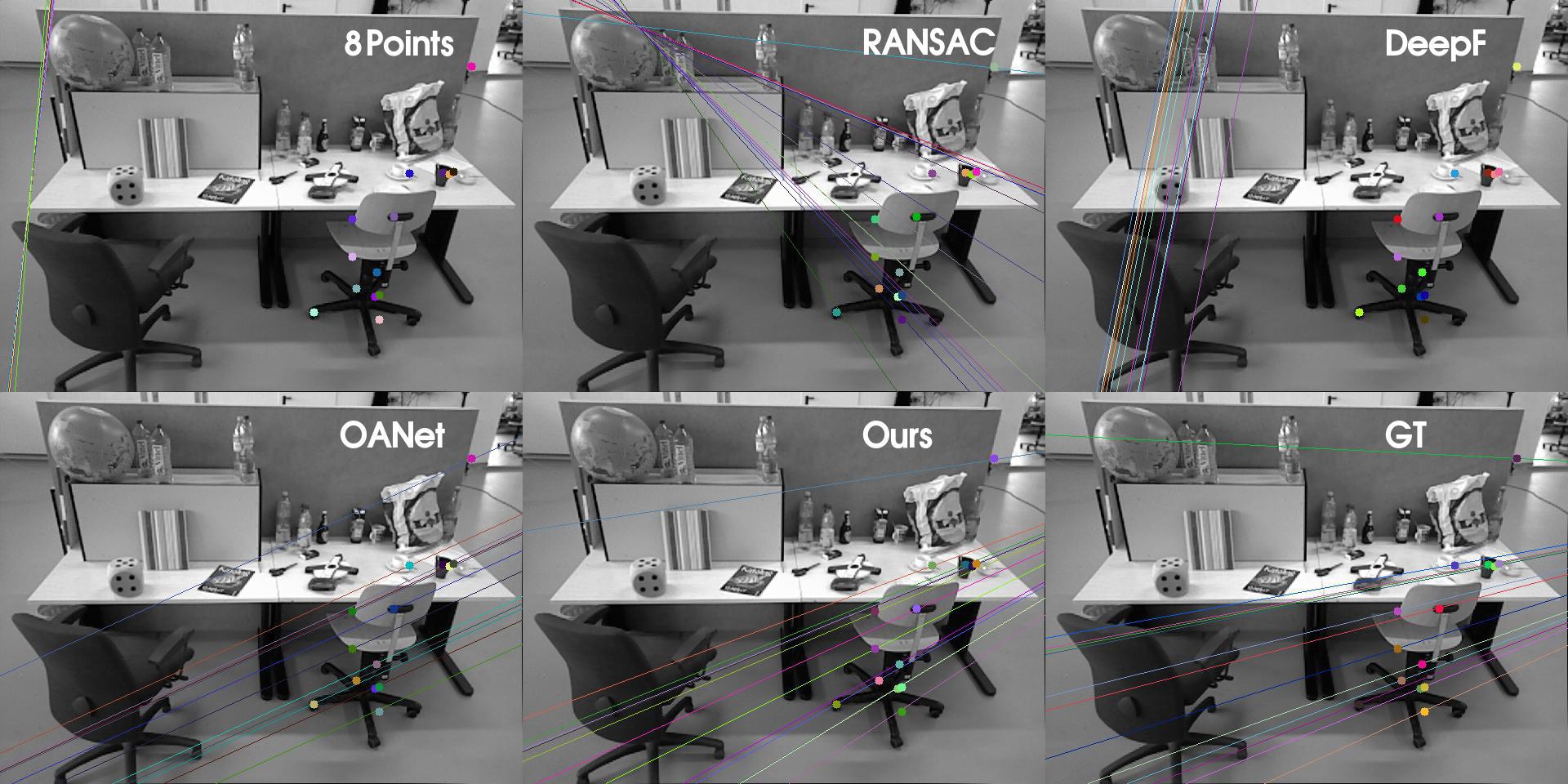}
	\caption{\small Visualization on TUM SLAM dataset. It can be seen that all the algorithms get worse results than those on KITTI dataset. It is possible that the big variation of fundamental matrix between image pairs and lack of image texture make the estimation a hard task. And our network gets the best result although the error regarding the epipolar lines orientation exists.}
	\label{Vis:T}
\end{figure}
       
And the visualization results are shown on Fig. \ref{Vis:K}-\ref{Vis:T}.

On KITTI dataset(Fig. \ref{Vis:K}), it can be seen that SIFT+8-Point algorithm get the worst performance, and the epipolar lines drawn by SIFT+RANSAC algorithm get most of points lied on them but their orientations have slight deviation. On the other hand, the visualization of DeepF shows some points which don't lie on correspond epipolar lines. And the visualization of OANet also has a little error regarding the epipolar lines orientation and our algorithm's visualization results are the closest to ground truth.

While on TUM SLAM dataset(Fig. \ref{Vis:T}), the difficulty of fundamental matrix estimation has greatly improved. The deviation of epipolar lines orientation means the wrong estimated model of algorithms and these orientation error are consistent of bad results on $M_{EA}$ in Tab. \ref{Res:table}. And our algorithm's visualization result is the closet to ground truth.

\subsection{Comprehensive Evaluation Metrics}
It can be seen in visualization results that errors of epipolar lines drawn by fundamental matrix are concentrated in two aspect: the relationship with correspond points and the orientation. The commonly used $M_{EC},M_{ED}$ metrics can measure the error of relationship between points and epipolar lines. And as the scores on these metrics get worse in Tab. \ref{Res:table}, the distance between points and correspond epipolar lines get greater on Fig. \ref{Vis:K}-\ref{Vis:T}. On the other hand, the $M_{EA}$ we proposed can indicate the error of epipolar lines orientation. In Tab. \ref{Res:table} SIFT+RANSAC algorithm get good scores on $M_{EC},M_{ED} $, but the orientation error in visualization results can only indicated by $M_{EA}$.

\subsection{Ablation Studies}
As mentioned before, our method has two main innovations: 1. the end-to-end network architecture with a completed pipeline. 2. the novel loss function including epipolar constraint. So we conducted ablation studies about them.

\subsubsection{Architecture with completed pipeline}
Our network architecture includes a completed pipeline for fundamental matrix estimation as traditional methods do. To demonstrate the efficacy of this architecture, we trained a model without epipolar constraint loss function which only includes the architecture on KITTI data-set and compared it with DeepF and OANet. The results are shown in Tab. \ref{Abl:Arch}.

	\begin{table}[t]
	  \centering
	  \caption{\small Ablation studies about network architecture with completed pipeline. We train a model without epipolar constraint loss function($l_e$) which only includes the architecture on KITTI dataset and compare it with DeepF and OANet.}
      \begin{tabular}{c|ccc} \hline
        Method & $M_{EC}$ & $M_{ED}$ & $M_{EA}$(${}^{\circ}$) \\ \hline 
        DeepF\cite{Poursaeed:2018vh} & 0.62 & 118.66 & 6.00 \\
        SIFT+OANet\cite{Zhang:2019ub}+RANSAC & 0.27 & 43.48  & 3.32 \\
        Ours without $l_e$ & $1.49$ & $7.158$ & $0.92$ \\ \hline
      \end{tabular}
            \label{Abl:Arch}
    \end{table}
	
It can be seen in Tab. \ref{Abl:Arch} that our network without $l_e$ get bad results on $M_{EC}$ but better scores on $M_{ED},M_{EA}$. That means this network architecture can improve the accuracy of fundamental matrix estimation in general.
	
\subsubsection{Epipolar constraint loss function}
A novel loss function which includes the L1, L2 loss function and epipolar constraint is proposed to guide our network training. And we add this epipolar constraint loss function($l_e$) on DeepF\cite{Poursaeed:2018vh} and remove it on our algorithm to see its improvement on fundamental matrix estimation algorithms. And the results are shown in Tab \ref{Abl:EL} which indicate that $l_e$ can improve the DeepF\cite{Poursaeed:2018vh} and our algorithm a lot on performance. 

	\begin{table}[t]
     \centering 
     \caption{\small Ablation studies about geometry constraint loss function. It improves the performance of DeepF and our network significantly. }
      \begin{tabular}{c|ccc} \hline
        Method & $M_{EC}$ & $M_{ED}$ & $M_{EA}$(${}^{\circ}$) \\ \hline
        DeepF\cite{Poursaeed:2018vh} & 0.62 & 118.66 & 6.00 \\
        DeepF\cite{Poursaeed:2018vh}+$l_e$ & $0.35\downarrow$ & $1.91\downarrow$ & $4.91\downarrow$ \\
        Ours without $l_e$ & $1.49$ & $7.158$ & $0.92$ \\ 
        Ours & $0.043\downarrow$ & $0.72\downarrow$ & $0.45\downarrow$ \\\hline
      \end{tabular}
            \label{Abl:EL}
    \end{table}  
   
  \subsection{Running speed}
  The running time of different algorithms are summarized in Tab. \ref{Ex:Ti}. Although our network doesn't have strength in running speed, it is faster than traditional methods.
	
	\begin{table}[t]
	\centering
	\caption{\small Running time of different fundamental matrix estimation algorithms. Our network doesn't have large advantage in running time, but it still takes less time than traditional methods.}
      \begin{tabular}{c|c} \hline
        Method & Time per pair(ms) \\ \hline
        SIFT+8Point & 437.25 \\
        SIFT+RANSAC & 452.28 \\
        DeepF & 159.75 \\
        SIFT+OANet+RANSAC & 490.45 \\ 
        Ours & 319.52 \\
        \hline
      \end{tabular}
      \label{Ex:Ti}
    \end{table}

\section{Conclusion}
In this paper we proposed a network architecture to perform end-to-end fundamental matrix estimation directly from stereo images. This network establishes a completed fundamental matrix estimation pipeline by different network architectures. And a novel loss function is introduced to keep the geometry property of fundamental matrix during estimation when training. We also design a novel metric to evaluate the estimation accuracy more comprehensively by measuring the error regarding epipolar lines orientation. Our experiments on both outdoor and indoor scenes show this network can estimate the fundamental matrix from stereo images directly and its precision exceeds traditional methods and some previous neuron network-based methods.

\bibliographystyle{ieeetr}
\bibliography{references.bib}

\begin{thebibliography}{10}

\bibitem{Luong:1996kc}
Q.-T. Luong and O.~D. Faugeras, ``{The fundamental matrix: Theory, algorithms,
  and stability analysis},'' {\em International Journal of Computer Vision},
  vol.~17, pp.~43--75, Jan. 1996.

\bibitem{Faugeras:1992dn}
O.~D. Faugeras, Q.~T. Luong, and S.~J. Maybank, ``{Camera self-calibration:
  Theory and experiments},'' in {\em Computer Vision {\textemdash} ECCV'92},
  pp.~321--334, Berlin, Heidelberg: Springer, Berlin, Heidelberg, May 1992.

\bibitem{Boukamcha:2019tj}
H.~Boukamcha, M.~Atri, and F.~Smach, ``{A real-time auto calibration technique
  for stereo camera},'' {\em International Journal of Computer Aided
  Engineering and Technology}, vol.~12, p.~74, Nov. 2019.

\bibitem{Mallon:bx}
J.~Mallon, P.~W.~I. Computing, Vision, and {2005}, ``{Projective rectification
  from the fundamental matrix},'' {\em Elsevier}.

\bibitem{Engineering:ck}
Q.~F. J. o. A.~S. Engineering, , and {2019}, ``{Efficient Fundamental Matrix
  Estimation for Robotic Visual Servoing Based on Continuous-time
  Optimization},'' {\em tku.edu.tw}.

\bibitem{international:fa}
D.~L. P. o. t. s.~I. international and {1999}, ``{Object recognition from local
  scale-invariant features},'' {\em ieeexplore.ieee.org}.

\bibitem{Fischler:1987bf}
M.~A. Fischler and R.~C. Bolles, {\em {Random Sample Consensus: A Paradigm for
  Model Fitting with Applications to Image Analysis and Automated
  Cartography}}.
\newblock Morgan Kaufmann Publishers, Inc., 1987.

\bibitem{Choy:2016up}
C.~B. Choy, J.~Gwak, S.~Savarese, and M.~Chandraker, ``{Universal
  Correspondence Network},'' {\em arXiv.org}, June 2016.

\bibitem{Poursaeed:2018vh}
O.~Poursaeed, G.~Yang, A.~Prakash, Q.~Fang, H.~Jiang, B.~Hariharan, and
  S.~Belongie, ``{Deep Fundamental Matrix Estimation without
  Correspondences},'' {\em arXiv.org}, Oct. 2018.

\bibitem{Yi:2017wj}
K.~M. Yi, E.~Trulls, Y.~Ono, V.~Lepetit, M.~Salzmann, and P.~Fua, ``{Learning
  to Find Good Correspondences},'' {\em arXiv.org}, Nov. 2017.

\bibitem{Zhang:2019ub}
J.~Zhang, D.~Sun, Z.~Luo, A.~Yao, L.~Zhou, T.~Shen, Y.~Chen, L.~Quan, and
  H.~Liao, ``{Learning Two-View Correspondences and Geometry Using Order-Aware
  Network},'' {\em arXiv.org}, Aug. 2019.

\bibitem{Qi:2017vq}
C.~R. Qi, H.~Su, K.~Mo, and L.~J. Guibas, ``{PointNet: Deep Learning on Point
  Sets for 3D Classification and Segmentation},'' pp.~652--660, 2017.

\bibitem{bian2019evaluation}
J.-W. Bian, Y.-H. Wu, J.~Zhao, Y.~Liu, L.~Zhang, M.-M. Cheng, and I.~Reid, ``An
  evaluation of feature matchers for fundamental matrix estimation,'' 2019.

\bibitem{Zhang:1998il}
Z.~Zhang, ``{Determining the Epipolar Geometry and its Uncertainty: A
  Review},'' {\em International Journal of Computer Vision}, vol.~27,
  pp.~161--195, Mar. 1998.

\bibitem{Armangue:uh}
X.~Armangu{\'e}, J.~S.~I. computing, vision, and {2003}, ``{Overall view
  regarding fundamental matrix estimation},'' {\em Elsevier}.

\bibitem{Lacey:vg}
A.~J. Lacey, N.~Pinitkarn, N.~T. BMVC, and {2000}, ``{An Evaluation of the
  Performance of RANSAC Algorithms for Stereo Camera Calibrarion.},'' {\em
  tina.wiau.man.ac.uk}.

\bibitem{Torr:2002jb}
P.~H.~S. Torr, ``{Bayesian Model Estimation and Selection for Epipolar Geometry
  and Generic Manifold Fitting},'' {\em International Journal of Computer
  Vision}, vol.~50, pp.~35--61, Oct. 2002.

\bibitem{Hartley:2003un}
R.~Hartley and A.~Zisserman, ``{Multiple view geometry in computer vision},''
  2003.

\bibitem{Eigen:2014vq}
D.~Eigen, C.~Puhrsch, and R.~Fergus, ``{Depth Map Prediction from a Single
  Image using a Multi-Scale Deep Network},'' {\em arXiv.org}, June 2014.

\bibitem{Melekhov:2017to}
I.~Melekhov, J.~Ylioinas, J.~Kannala, and E.~Rahtu, ``{Relative Camera Pose
  Estimation Using Convolutional Neural Networks},'' {\em arXiv.org}, Feb.
  2017.

\bibitem{Ummenhofer:2017uq}
B.~Ummenhofer, H.~Zhou, J.~Uhrig, N.~Mayer, E.~Ilg, A.~Dosovitskiy, and
  T.~Brox, ``{DeMoN: Depth and Motion Network for Learning Monocular Stereo},''
  pp.~5038--5047, 2017.

\bibitem{DeTone:vr}
D.~DeTone, T.~Malisiewicz, A.~R. a. p.~a. 1606.03798, and {2016}, ``{Deep image
  homography estimation},'' {\em arxiv.org}.

\bibitem{Nguyen:gw}
T.~Nguyen, S.~W. Chen, S.~S. I.~R. and, and {2018}, ``{Unsupervised deep
  homography: A fast and robust homography estimation model},'' {\em
  ieeexplore.ieee.org}.

\bibitem{:ue}
S.~A. a. p.~a. 1709.03524 and {2017}, ``{Recovering homography from camera
  captured documents using convolutional neural networks},'' {\em arxiv.org}.

\bibitem{DeTone:2018tv}
D.~DeTone, T.~M. P.~o. the, and {2018}, ``{Superpoint: Self-supervised interest
  point detection and description},'' {\em openaccess.thecvf.com}.

\bibitem{DeTone:2018tp}
D.~DeTone, T.~Malisiewicz, A.~R. a. p.~a. 1812.03245, and {2018},
  ``{Self-improving visual odometry},'' {\em arxiv.org}.

\bibitem{Luo:2018wq}
Z.~Luo, T.~Shen, L.~Zhou, S.~Zhu, R.~Zhang, Y.~Yao, T.~Fang, and L.~Quan,
  ``{GeoDesc: Learning Local Descriptors by Integrating Geometry
  Constraints},'' {\em arXiv.org}, pp.~170--185, July 2018.

\bibitem{Luo:2019ub}
Z.~Luo, T.~Shen, L.~Zhou, J.~Z. P.~o. the, and {2019}, ``{Contextdesc: Local
  descriptor augmentation with cross-modality context},'' {\em
  openaccess.thecvf.com}.

\bibitem{Li:ie}
R.~Li, S.~Wang, Z.~Long, D.~G. .~I. international, and {2018}, ``{Undeepvo:
  Monocular visual odometry through unsupervised deep learning},'' {\em
  ieeexplore.ieee.org}.

\bibitem{Rocco:uc}
I.~Rocco, R.~Arandjelovic, J.~S. P. o.~t. IEEE, and {2017}, ``{Convolutional
  neural network architecture for geometric matching},'' {\em
  openaccess.thecvf.com}.

\bibitem{Ranftl:2018wc}
R.~Ranftl and V.~Koltun, ``Deep fundamental matrix estimation,'' in {\em
  Computer Vision -- ECCV 2018} (V.~Ferrari, M.~Hebert, C.~Sminchisescu, and
  Y.~Weiss, eds.), (Cham), pp.~292--309, Springer International Publishing,
  2018.

\bibitem{He:tt}
K.~He, X.~Zhang, S.~Ren, J.~S. recognition, pattern, and {2016}, ``{Deep
  residual learning for image recognition},'' {\em openaccess.thecvf.com}.

\bibitem{Geiger:2013kp}
A.~Geiger, P.~Lenz, C.~Stiller, and R.~Urtasun, ``{Vision meets robotics: The
  KITTI dataset:},'' {\em The International Journal of Robotics Research},
  vol.~32, pp.~1231--1237, Aug. 2013.

\end{thebibliography}

\end{document}